\documentclass[runningheads]{llncs}
\usepackage{graphicx}

\usepackage{tikz}
\usepackage{comment}
\usepackage{amsmath,amssymb} 
\usepackage{color}
\usepackage{orcidlink}
\usepackage{booktabs}
\usepackage{amsfonts}
\usepackage[product-units=repeat]{siunitx}
\usepackage{float}
\usepackage{subcaption}
\usepackage{multirow}

\usepackage[accsupp]{axessibility}  

\begin{document}
\pagestyle{headings}
\mainmatter
\def\ECCVSubNumber{3604}  

\title{TREND: Truncated Generalized Normal Density Estimation of Inception Embeddings\\ for GAN Evaluation} 

\titlerunning{TREND for GAN Evaluation}

\author{Junghyuk Lee\orcidlink{0000-0002-6164-0728} \and
Jong-Seok Lee\orcidlink{0000-0002-8038-1119}}
\authorrunning{J. Lee and J.-S. Lee.}
%
\institute{School of Integrated Technology, Yonsei University, Seoul, Republic of Korea\\
\email{\{junghyuklee,jong-seok.lee\}@yonsei.ac.kr}}
%

\maketitle

\begin{abstract}
Evaluating image generation models such as generative adversarial networks (GANs) is a challenging problem. A common approach is to compare the distributions of the set of ground truth images and the set of generated test images. The Frech\'et Inception distance is one of the most widely used metrics for evaluation of GANs, which assumes that the features from a trained Inception model for a set of images follow a normal distribution. In this paper, we argue that this is an over-simplified assumption, which may lead to unreliable evaluation results, and more accurate density estimation can be achieved using a truncated generalized normal distribution. Based on this, we propose a novel metric for accurate evaluation of GANs, named TREND (TRuncated gEneralized Normal Density estimation of inception embeddings). We demonstrate that our approach significantly reduces errors of density estimation, which consequently eliminates the risk of faulty evaluation results. Furthermore, the proposed metric significantly improves robustness of evaluation results against variation of the number of image samples.
\end{abstract}

\section{Introduction}
Generative models for realistic image generation is one of the most active research topics in recent days. The objective of the generative models is to find a mapping from random noise to real images by estimating probability density $P_g$ from target distribution $P_r$. Among different types of generative models, generative adversarial networks (GANs) are particularly popular, which learn the target distribution by solving the objective equation $P_g = P_r$ as a min-max game of a generator and a discriminator~\cite{goodfellow2014generative}. Recent state-of-the-art GANs~\cite{biggan,sg2,projgan} can generate highly realistic images such as faces, animals, structures, etc.

Evaluation of GAN models is crucial for developing models and improving their performance. Assessing the quality of generated images via subjective tests is inadequate due to the issues of excessive time and cost. Accordingly, performance evaluation is usually based on measuring the likelihood of the learned probability density $P_g$ with respect to the ground truth $P_r$. However, since $P_g$ defined by GANs is implicit, it is difficult to directly measure the likelihood. Therefore, evaluation of GANs is usually based on sample statistics to estimate $P_g$ and $P_r$ for comparison.

Building the distribution from generated or real image samples is a challenging part in GAN evaluation. In early literature, there exist attempts to directly measure likelihood using a kernel density estimation method. However, due to high dimensionality of pixel-domain images, this method requires a substantial number of samples. Moreover, it is noted that the measured likelihood is sometimes unrelated to the quality of generated images~\cite{theis2016note}.

In order to address the high dimensionality and sample quality issues, the Inception score (IS)~\cite{IS} proposes to use an Inception model that is trained for image classification~\cite{Inception}. It measures the Kullback-Leibler divergence (KLD) of the conditional label distribution for generated images and the marginal distribution of the pre-trained model. Although IS performs well, it has major drawbacks as well. It measures correctness of generated images compared to the classification model, instead of considering the target distribution of GANs, which causes inability to detect overfitting and mode collapse~\cite{empirical}.

The Fr\'echet Inception distance (FID)~\cite{FID} also uses a pre-trained Inception model but in a different way from IS. It uses the output of a specific layer of the Inception model, called Inception feature, to embed sampled images to an informative domain. Then, the Fr\'echet distance, also known as earth mover’s distance, is measured between the Inception features of generated test samples and those of target real samples. Showing better performance than other metrics, FID is one of the most frequently used metrics for evaluation of GANs nowadays.

Despite its widespread usage, we argue that FID has several drawbacks. As a major drawback, we find out that FID incorrectly estimates the distribution of Inception features. FID assumes that Inception features follow a normal distribution, which is not accurate for real data. First, the distribution of Inception features is truncated at zero due to the rectified linear unit (ReLU) applied to obtain the features, which is also noted in~\cite{KID}. Second, the shape of the distribution is significantly different from the normal distribution, having a sharper peak. In addition, FID has a high bias in terms of the number of samples. Although a method reducing the bias is proposed in~\cite{infinity}, it is still based on FID under the normality assumption.

In this paper, we propose a novel method for accurate GAN evaluation, which is named TREND (TRuncated gEneralized Normal Density estimation of inception embeddings). In order to address the aforementioned issues, we thoroughly analyze Inception features with respect to their distirubional properties. We find that the truncated generalized normal distribution can effectively model the probability density of Inception features of real-world images, based on which we design the proposed TREND metric. Our main contributions are as follows: 
\begin{itemize}
    \item We analyse Inception features and show that density estimation using the normal distribution is inaccurate in conventional evaluation methods. We conduct thorough and complete analysis regarding the distribution of Inception embeddings.
    \item Based on the analysis, we propose to model the distribution of Inception features with the truncated generalized normal distribution and measure the Jensen-Shannon divergence between the estimated distributions of generated and real images.
    \item We demonstrate that the proposed method can accurately evaluate various generative models including not only GANs but also variational autoencoders (VAEs) and diffusion models compared to existing metrics. Furthermore, we show that the proposed method removes the bias caused by the variation of the number of samples.
\end{itemize}

\section{Related Work} \label{sec:related}
\subsection{GANs}
Generative models aim to capture the probability distribution of target real image data, $P_r$. After training, one can generate new data according to the learned probability density $P_g$. Among generative models, GANs~\cite{goodfellow2014generative} train a generator ($G$) and a discriminator ($D$) playing a min-max game to find a Nash equilibrium:
\begin{equation}
    \min_G \max_D \mathbb{E}_{x^r\sim P_r}[\log D(x^r)]+\mathbb{E}_{z\sim P_z}[\log(1-D(G(z)))],
\end{equation}
where $x^r$ is a sample from the target distribution $P_r$ and $z$ is a latent vector drawn from the latent distribution $P_z$ that is usually set to be a normal or uniform distribution.

Plenty of studies on image generation using GANs have been conducted with variations such as modification of the loss function~\cite{wgan}, model architecture~\cite{dcgan,sagan}, normalization strategy~\cite{sngan}, and up-scaling approach~\cite{progan,sg,sg2,biggan} in order to improve stability of learning and to enhance the quality and resolution of generated images.
Popular GAN models include DCGAN~\cite{dcgan}, ProGAN~\cite{progan}, StyleGAN~\cite{sg}, StyleGAN2~\cite{sg2}, and BigGAN~\cite{biggan}.

\subsection{Evaluation Metrics for GANs}
A common procedure for GAN evaluation is composed of the following steps. The first step is to prepare a set of generated images from the test GAN model and a set of real images from the target dataset (e.g., ImageNet). Second, an embedding function is applied to extract low-dimensional informative features from the images (e.g., Inception feature embedding). Then, the probability density of each set of features is estimated for comparison. Finally, difference of the two distributions is measured using a proper metric, where a smaller difference indicates better performance of the GAN. 

IS~\cite{IS} uses the pre-trained Inception model for both embedding and density estimation. Given test images, it measures the KLD of the conditional probability $p(y|x)$ from the marginal distribution $p(y)$ using the softmax output of the Inception model: 
\begin{equation}
    IS = \exp \left(\mathbb{E}_{x^g} [KLD(p(y|x^g)\|p(y))]\right),
\end{equation}
where $x^g$ is a test data (i.e., generated image) and $y$ is the predicted class label.
Since IS does not consider the target distribution and only uses the conditional probability estimated by the Inception model for the generated images, its adequacy has been controversial~\cite{empirical}. For example, it favors highly classifiable images instead of high quality images.

FID~\cite{FID} also uses the Inception model for image embedding. Assuming that the distribution of embedded features is Gaussian, it measures the Fr\'echet distance between the Gaussian distributions for the generated and real data, i.e., 
\begin{equation} \label{eq:FID}
    FID = \left\|\mu^{g}-\mu^{r}\right\|^2_2 + \mathrm{Trace}\left(\Sigma^g+\Sigma^r-2\left(\Sigma^g\Sigma^r\right)^{1/2}\right), 
\end{equation}
where $\left\|\cdot\right\|^2_2$ denotes the $l2$-norm operator and $\left(\mu^g,\Sigma^g\right)$ and $\left(\mu^r,\Sigma^r\right)$ are the mean and the covariance of the generated and real data, respectively.
Having a straightforward approach and formula, it is commonly used for GAN evaluation in recent days. 
However, it has been argued that FID is biased~\cite{infinity,fidself} and the normality is not guaranteed~\cite{KID}.
In order to address the bias problem, extrapolating FID with respect to the number of samples is proposed in~\cite{infinity}. Nonetheless, the issue of inaccurate normal density estimation still remains.
In addition, FID is unexpectedly susceptible to low-level preprocessing such as resizing and compression~\cite{aliased}.

The Kernel Inception distance (KID)~\cite{KID} measures the maximum mean discrepancy of the two distributions after transforming the Inception features using a kernel function. While KID estimates feature distributions without normality assumption, the choice of a proper kernel function has not been well studied.

Measuring different aspects (e.g., fidelity and diversity) of generated images separately has been also considered~\cite{PR,IPR,DC}. 
For instance, the improved precision and recall method~\cite{IPR} applies the precision and recall approach in machine learning to real and generated images for GAN evaluation. 
Although such an approach can be effective for a diagnostic purpose, using a single-valued metric facilitates more efficient and convenient evaluation and comparison of GAN models, and thus has been more popular.

\section{Analysis of Inception Features} \label{sec:analysis}
The Inception model pre-trained using the ImageNet dataset~\cite{imagenet} is widely used as an image embedding function in most state-of-the-art GAN evaluation metrics~\cite{IS,FID,KID,infinity}. The 2048-dimensional Inception feature is the output of the last pooling layer before the fully connected layer of the model. In this section, we thoroughly analyze the distribution and characteristics of the Inception features. 

\begin{figure}[t]
\centering
\includegraphics[width=1\linewidth]{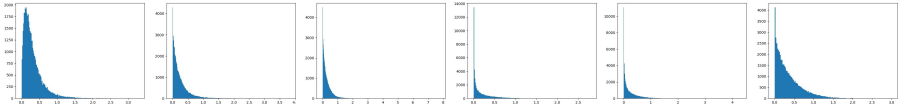}
\caption{Histograms of the Inception features from the ImageNet validation dataset for selected dimensions. In each figure, the x-axis and y-axis are the feature value and the frequency, respectively.}
\label{fig:acts}
\end{figure}

In \figurename~\ref{fig:acts}, histograms of the Inception features from the validation split of the ImageNet dataset are shown. Some representative feature dimensions are chosen out of the 2048 dimensions.
From the figure, we find the following observations. 

First, the distribution of the Inception feature is left-truncated. This is because the Inception model uses the ReLU as the activation function, by which the negative values are set to zero. 

Second, the shapes of the distributions differ from normal distributions. The distributions in~\figurename~\ref{fig:acts} have sharper peaks than normal distributions, i.e., they are leptokurtic. When the truncation at zero is excluded, the kurtosis value of the Inception features for the ImageNet dataset is measured as 28.6 on average across dimensions, which is larger than that of the normal distribution (which is 3). Furthermore, the measured kurtosis ranges from 1.3 to 592.2 with a median of 8.5, implying that the sharpness significantly varies according to the dimension.

Third, the Inception feature dimensions are nearly independent with each other. 
We measure the Pearson correlation coefficient (PCC) value between each pair of dimensions. 
The average PCC is 0.055 with a standard deviation of 0.046.
While FID uses a multivariate normal distribution, the independence allows us to separately estimate the dimension-wise distributions.
This can significantly reduce the number of parameters to be estimated.
More details of the independence are presented in Section~\ref{sup:independence}.

FID does not consider these characteristics, which consequently may lead to unreliable and inaccurate evaluation results. In order to address this issue, we propose a new method with more accurate modeling of the distributions of the Inception features in the following section.

\section{Proposed Method} \label{sec:proposed}
A brief summary of the proposed method called TREND is as follows. First, we extract the $d$-dimensional Inception feature $x \in \mathbb{R}^{d}$ from an image (i.e., $d=2048$).
Next, we model the probability density of the Inception feature as \begin{equation}
f(x)  = \textrm{TGN}(x|M,S,B)    ,
\end{equation}
where TGN is a multivariate truncated generalized normal distribution with mean $M$, covariance $S$, and shape parameter $B$. 
This is much more flexible than the normal distribution, allowing us to model the truncation at zero and peak sharpness varying with respect to the feature dimension. Finally, we use the Jensen-Shannon divergence (JSD) as a dissimilarity metric between the estimated test and target distributions.

Based on the observation in Section~\ref{sec:analysis}, we assume independence between feature dimensions. Therefore, we can replace the multivariate probability density with a product of dimension-wise distributions:
\begin{equation}
    f(x)=\prod_{i=1}^{d}f_i(x_i),
\end{equation}
where $x_i$ is the $i$th dimension of $x$ and $f_i(x_i)$ is a one-dimensional truncated generalized normal distribution for the $i$th dimension:
\begin{equation} \label{eq:TGD}
    f_i(x_i)=\frac{\beta}{\sigma G}{e^{-{\left|\frac{x_i-\mu}{\sigma}\right|}^{\beta}}},
\end{equation}
where $\mu$, $\sigma$, and $\beta$ are the mean, standard deviation, and shape parameter, respectively\footnote{We omit subscript $i$ for $\mu$, $\sigma$, $\beta$, and $G$ for simplicity.}. Note that $\beta<2$ for leptokurtic distributions. $G$ is a scale factor for normalization due to truncation, which is defined as
\begin{equation}\label{eq:g}
    G = \gamma\left(\frac{1}{\beta},{\left|\frac{A_1-\mu}{\sigma}\right|}^{\beta}\right)+\gamma\left(\frac{1}{\beta},{\left|\frac{A_2-\mu}{\sigma}\right|}^{\beta}\right),
\end{equation} 
where $A_1$ and $A_2$ are the lower and upper truncation points, respectively. $\gamma(u,v)$ is the lower incomplete gamma function with upper limit $v$ of integral, i.e., 
\[
\gamma(u,v)=\int^v_0 t^{u-1}e^{-t}\,dt.
\]
We set lower and upper truncation points (i.e., $A_1$ and $A_2$) to zero and positive infinity, respectively.
Then, \eqref{eq:g} becomes
\begin{equation}
    G = \gamma\left(\frac{1}{\beta},{\left|\frac{\mu}{\sigma}\right|}^{\beta}\right)+\Gamma\left(\frac{1}{\beta}\right).
\end{equation}

The parameters $(\mu, \sigma, \beta)$ are estimated by maximizing the likelihood of $n$ samples from the distribution. When we denote the $i$th feature dimension of the $j$th sample as $x^j_i$~$(j=1,\ldots,n)$, the likelihood $L$ is written as
\begin{equation} \label{eq:l}
    L = \left(\frac{\beta}{\sigma G}\right)^n \prod_{j=1}^n e^{-{\left|\frac{x_i^j-\mu}{\sigma}\right|}^{\beta}} .
\end{equation} 
By taking logarithm, we get log-likelihood as follows:
\begin{equation} \label{eq:ll}
    \log L = n \log \beta-n\log\sigma- n\log G-\sum_{j=1}^{n} {\left|\frac{x_i^j-\mu}{\sigma}\right|}^{\beta}.
\end{equation}

Since it is intractable to formulate an analytic solution for the parameters $(\mu,\sigma,\beta)$ maximizing \eqref{eq:ll}, a numerical approach should be adopted to estimate the parameters.
We use a trust-region minimization method~\cite{optimizer1998}.
For the density estimation, we omit the feature values at zero occurring by truncation.

Finally, we measure the JSD between the estimated probability density of the Inception features for the generated images $f^g$ and that for the target real images $f^r$ as 
\begin{equation} \label{eq:trend}
    TREND = JSD(f^g,f^r).
\end{equation}
JSD measures the divergence of each distribution from their average distribution using KLD and is defined by
$
    JSD(p,q) = \frac{1}{2}\left(KLD(p\|m)+KLD(q\|m)\right),
$
where $m=(p+q)/2$.
Since we assume independence between feature dimensions, \eqref{eq:trend} can be written as the average of the dimension-wise JSDs:
\begin{equation}
    TREND(f^g,f^r) = \frac{1}{d}\sum_{i=1}^{d}{JSD(f^g_i,f^r_i)}.
\end{equation}

There are two reasons of using JSD instead of the Frech\'et distance used in FID to compare distributions.
First, it is too complex to calculate the Frech\'et distance between general distributions other than Gaussian distributions. Furthermore, JSD is bounded within $[0,1]$ and thus more intuitive to interpret the result of performance comparison than the Fr\'echet distance that has only the lower bound of 0.
TREND reaches its minimum value of 0 when the test and target distributions are identical, which can be achieved for an ideal GAN.
Conversely, when the distributions are completely different from each other, TREND yields its maximum value of 1.

\section{Experiments} \label{sec:experiments}
In this section, we conduct various experiments to demonstrate that the proposed method enables accurate and effective performance evaluation of GANs.
Details of the experimental setup can be found in Section~\ref{sup:setup}.

\subsection{Choice of Distribution}

\begin{figure}[t]
\centering
\begin{minipage}[t]{.58\textwidth}
\subfloat[]{\includegraphics[width=0.45\linewidth]{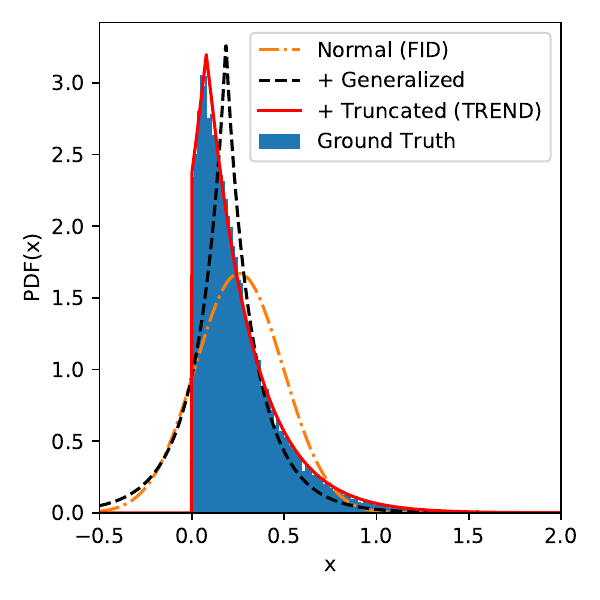}%
\label{fig:choice1}
}
\subfloat[]{\includegraphics[width=0.45\linewidth]{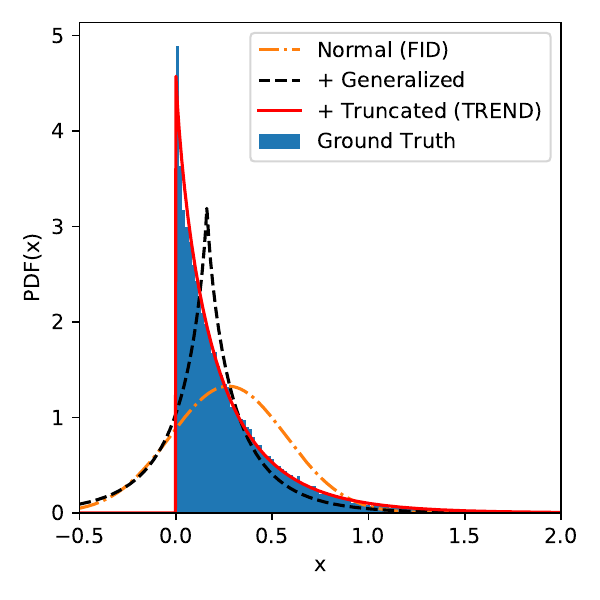}%
\label{fig:choice2}}
\caption{Ground truth histograms and estimated densities of the Inception features in (a) the 120th dimension and (b) the 504th dimension. The estimated parameters $(\mu, \sigma, \beta)$ by TREND are $(0.08, 0.25, 1.03)$ and $(-1.9\times 10^{-11}, 0.19, 0.82)$, respectively.
}
\label{fig:choice}
\end{minipage}~~~%
\begin{minipage}[t]{0.38\textwidth}
\includegraphics[width=1\linewidth]{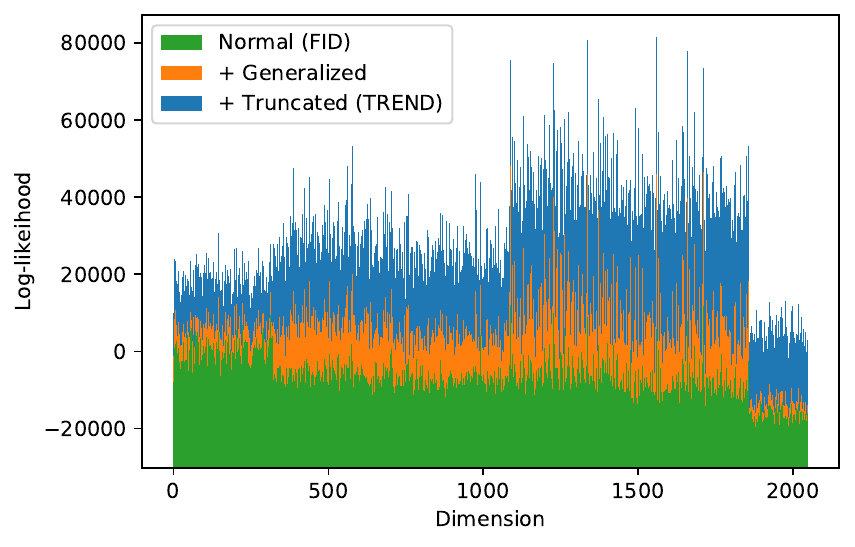}
\caption{Log-likelihoods from the estimated densities for each dimension of the Inception features.}
\label{fig:likelihood}
\end{minipage}
\end{figure}

In order to demonstrate that TREND can effectively estimate distributions of Inception features, we perform an ablation study with respect to the choice of distribution. We compare the normal distribution (as in FID), generalized normal distribution without truncation, and truncated generalized normal distribution used in the proposed TREND method for estimating the distribution of the Inception features for the ImageNet dataset. 

\figurename~\ref{fig:choice} shows the ground truth histograms of the Inception features at specific dimensions for all images as blue-colored bars. And, each line represents the estimated probability density using one of the distributions. Overall, the estimated densities using the truncated generalized normal distribution used in TREND conform best to the ground truth distributions. On the other hand, in the cases of the normal distribution, the estimated densities significantly deviate from the ground truth on both left and right sides. Not only the shapes, but also the peak locations are far different from those of the ground truth. 
For the generalized normal distribution, in \figurename~\ref{fig:choice1}, the sharpness of the peak is estimated better than the normal distribution. However, due to the truncated region, it fails to estimate the peak location. In \figurename~\ref{fig:choice2}, inaccuracy of the generalized normal distribution is more prominent since truncation removes all the left tail and even some of the right tail near the peak.

For quantitative analysis, \figurename~\ref{fig:likelihood} shows log-likelihoods of the Inception features from the estimated densities using the three fitting distributions. The average log-likelihoods are -9463, 879, and 23867 for the normal distribution (FID), generalized normal distribution, and the truncated generalized normal distribution (TREND), respectively. As shown in \figurename~\ref{fig:likelihood}, TREND performs best with the highest likelihood values than the others in all feature dimensions. 
We also conduct one-tailed $t$-tests under the null hypothesis that the average log-likelihoods are the same between the truncated generalized normal distribution and one of the other two, which confirm the significance of the differences: $t(2047)=110$, $p<5\times10^{-16}$ for truncated generalized normal vs. normal; $t(2047)=69$, $p<5\times10^{-16}$ for truncated generalized normal vs. generalized normal. In conclusion, the distribution used in TREND is more appropriate to estimate the density of the Inception features than the other ones.

\subsection{Comparing Metrics using Toy Datasets}

In the previous experiments, the accuracies of density estimation using different distributions were compared. In this section, we investigate how the estimation accuracy affects the result of performance evaluation of GANs. In order to effectively demonstrate this, we build toy datasets using continuous distributions as if they are probability densities of the Inception features. We present two scenarios where FID fails to accurately determine the difference of distributions, while TREND does not.

\begin{figure}[t]
\centering
\begin{minipage}[t]{0.43\textwidth}
\includegraphics[width=\linewidth]{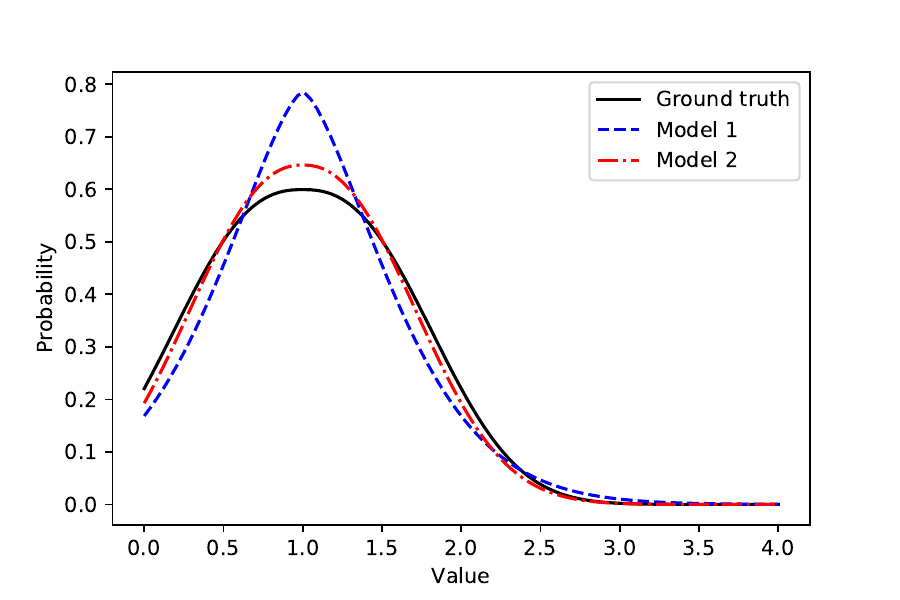}%
\caption{Probability densities for the first toy scenario. FIDs are \textbf{0.0017} (model 1) and 0.0018 (model 2), and TRENDs are 0.0086 (model 1) and \textbf{0.0022} (model 2).}
\label{fig:toy1}

\end{minipage}~~~~~~~~%
\begin{minipage}[t]{0.43\textwidth}
\includegraphics[width=\linewidth]{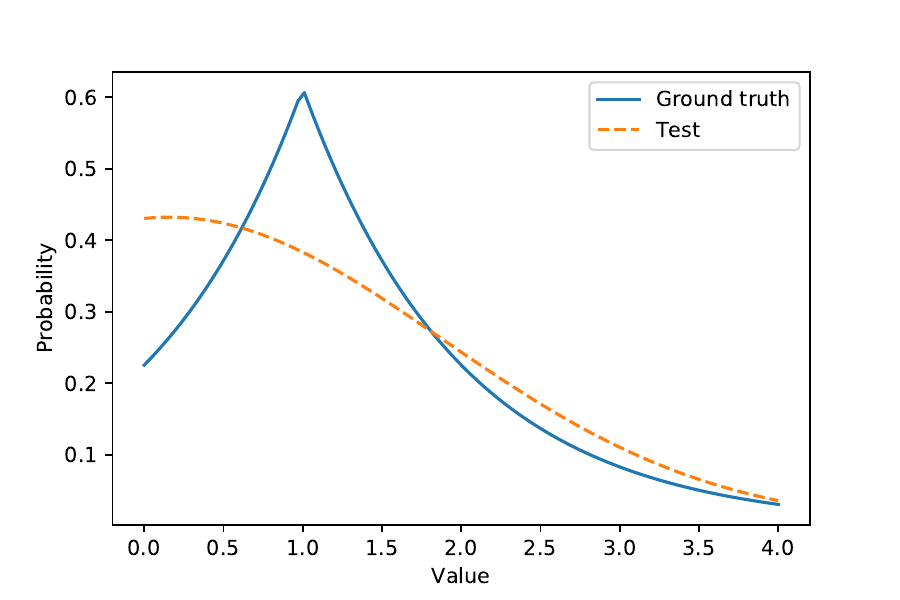}%
\caption{Probability densities for the second toy scenario.}
\label{fig:toy2}
\end{minipage}
\end{figure}

In the first scenario, we set a hypothetical density for ground truth images to a truncated generalized normal distribution. Two hypothetical GAN models (model 1 and model 2) are evaluated against the ground truth distribution, which are also modeled as truncated generalized normal distributions. The three distributions have the same $\mu$ but different $\sigma$ and $\beta<2$. 50000 random samples are drawn from each distribution, which correspond to the Inception features. The samples from model 1 and model 2 are evaluated against 50000 samples from the ground truth distribution using FID and TREND. When the distributions are compared in~\figurename~\ref{fig:toy1}, an accurate metric will determine that model 2 is a better approximation of the ground truth than model 1.
However, FID yields almost the same scores and even favors model 1 against model 2. On the contrary, the result of TREND is consistent with the expectation.

The second scenario considers a case where a GAN model fails to learn the ground truth distribution, as shown in~\figurename~\ref{fig:toy2}. Again, truncated generalized normal distributions are used for the ground truth distribution and the learned distribution by the GAN, from each of which 50000 samples are drawn. 
For TREND, the difference between the distributions is well captured with a score of 0.015. However, FID yields a score of $1.4\times10^{-4}$, determining that the difference is insignificant. 

In both scenarios, the failure of FID is due to inaccurate density estimation of non-Gaussian distributions using normal distributions. On the other hand, TREND provides more accurate density estimation and thus more accurate evaluation results.

\subsection{Density Estimation of Real-world Datasets} \label{sec:realdensity}

\begin{table}[t]
  \centering
  \caption{Estimated parameters for various datasets. In each case, the mean and standard deviation are shown.}
  \label{tab:param}
  \setlength\tabcolsep{10pt}
  \scriptsize
  \begin{tabular}{l|ccc}
    \toprule
    Dataset & $\mu$ & $\sigma$ & $\beta$\\
    \midrule
    CIFAR10 & 0.07$\pm$0.18 & 0.28$\pm$0.25 & 0.94$\pm$0.38\\
    DCGAN & 0.08$\pm$0.19 & 0.30$\pm$0.26 & 1.01$\pm$0.40\\
    \midrule
    ImageNet & 0.02$\pm$0.04 & 0.13$\pm$0.12 & 0.68$\pm$0.25\\
    BigGAN & 0.02$\pm$0.04 & 0.14$\pm$0.12 & 0.71$\pm$0.27\\
    \midrule
    CelebA & 0.10$\pm$0.20 & 0.26$\pm$0.24 & 1.04$\pm$0.45\\
    ProGAN & 0.14$\pm$0.22 & 0.25$\pm$0.21 & 1.11$\pm$0.43\\
    \midrule
    FFHQ & 0.09$\pm$0.16 & 0.28$\pm$0.23 & 1.04$\pm$0.39\\
    StyleGAN & 0.10$\pm$0.17 & 0.28$\pm$0.22 & 1.06$\pm$0.39\\
    StyleGAN2 & 0.09$\pm$0.17 & 0.28$\pm$0.23 & 1.05$\pm$0.41\\
  \bottomrule
\end{tabular}
\end{table}

In this section, we examine the distributions of Inception features of real-world image datasets. We estimate the parameters of the truncated generalized normal distribution as in TREND in each dimension of the Inception features and compare them for various datasets. We consider sets of generated images by DCGAN, ProGAN, BigGAN, StyleGAN, and StyleGAN2, and the original datasets (CIFAR10~\cite{cifar10}, CelebA~\cite{celeba}, ImageNet~\cite{imagenet}, and FFHQ~\cite{sg}) used to train the models.
\tablename~\ref{tab:param} shows the average values of the estimated parameters along with their standard deviations. 
Histograms of the estimated parameters are provided in \figurename~\ref{fig:hp}.

Since the images of different datasets have different characteristics, the estimated parameters are also different for each dataset. 
In the perspective of GAN evaluation, the estimated parameters for the images generated by BigGAN are similar to those for ImageNet.
On the other hand, the estimated parameters of DCGAN images are slightly different from those of the CIFAR10 dataset, which explains the well-known performance inferiority of DCGAN to BigGAN.

In terms of the shapes of the distributions, the average of shape parameter $\beta$ is smaller than 2 for all datasets, which indicates that the distributions are sharper than the normal distribution. Especially in the case of the ImageNet dataset, the distributions in all dimensions are sharper than the normal distribution. 
For the other datasets, only small numbers of dimensions have $\beta$ larger than 2. 

In terms of the location of the distribution, $\mu$ is slightly greater than zero on average. Nonetheless, 23.8\%, 22.6\%, 40.0\%, and 20.1\% of the feature dimensions have $\mu<0$ for CIFAR10, CelebA, ImageNet, and FFHQ, respectively, for which more than a half of a (non-truncated) generalized normal distribution is cut off. In these cases, the discrepancy from normal distributions becomes large as shown in~\figurename~\ref{fig:choice2}, thus density estimation and subsequent evaluation results would be particularly unreliable. 

\subsection{Effectiveness for Image Disturbance} \label{sec:disturbance}
For a metric performing GAN evaluation, it is important to be able to identify unnatural images that are from different distributions from those of natural images. In this section, we demonstrate that TREND can effectively capture the differences. 
We apply disturbances such as noise, where the difference can be easily determined by human judgement.
For the experiment, two sets of images are used, where one consists of the original images and the other consists of images with disturbance. Then, we measure the differences between two sets using FID and IS as well as TREND for comparison.

We use 50000 images from the ImageNet validation dataset.
For image disturbance, Gaussian noise with variance $\sigma_{gn}^2$, Gaussian blur with width $\sigma_{gb}$, and random erasing with erasing ratio $r$ are applied to the images.
We set three levels of disturbances: $\sigma_{gn}^2 \in \{0.05, 0.10, 0.15\}$, $\sigma_{gb} \in \{1,2,3\}$, and $r \in \{0.25, 0.50, 0.75\}$.
In general, the larger the disturbance level is, the larger the image differences are, and the larger the perceptual differences are.
An example of disturbed images is shown in \figurename~\ref{fig:manipulation}.

\begin{table}[t]
\centering
\caption{Evaluation results of disturbed images}
\label{tab:manipulation}
\scriptsize
\begin{tabular}{c|c|ccc}
\toprule
Evaluation metric & Disturbance level & Gaussian noise & Gaussian blur & Random erasing \\
\midrule
\multirow{3}{*}{IS}    & 1                 & 189.5          & 139.5         & 66.3           \\
                       & 2                 & 145.7          & 54.0          & 11.3           \\
                       & 3                 & 106.9          & 22.0          & 3.1           \\
\midrule
\multirow{3}{*}{FID}   & 1                 & 7.95           & 7.81          & 44.27          \\
                       & 2                 & 21.34          & 23.13         & 96.47          \\
                       & 3                 & 41.09          & 40.67         & 127.24         \\
\midrule
\multirow{3}{*}{TREND} & 1                 & 0.0058         & 0.0057        & 0.0369         \\
                       & 2                 & 0.0120         & 0.0127        & 0.0670         \\
                       & 3                 & 0.0208         & 0.0211        & 0.0827         \\
\bottomrule
\end{tabular}
\end{table}

The results are shown in \tablename~\ref{tab:manipulation}. 
In all cases, TREND becomes larger when the disturbance level increases, which demonstrates that TREND can effectively capture the deviations in distribution for the disturbed images.
FID behaves similarly to TREND. In the case of IS, the random erasing with level 1 is determined to be better than the Gaussian blur with level 2 (66.3 vs. 54.0), which is not consistent with human judgment.

\subsection{Evaluating Generative Models} \label{sec:benchmark}

In this section, we evaluate the performance of TREND for evaluation of state-of-the-art GANs.
Various types of GANs have been proposed so far, and their training datasets are also diverse, such as objects, structures, animals, and human faces. 
A metric for GANs should work well across the variety of training datasets of GANs.
Thus, we examine evaluation results of GANs that are trained on different datasets.
For comparison, we use not only IS, FID, and KID but also the improved precision and recall~\cite{IPR}, which is a two-dimensional metric.
We employ four GANs: DCGAN, BigGAN, StyleGAN, and StyleGAN2 and use 50000 generated images for each model to evaluate performance of the models. 
We also evaluate other generative models based on VAE and diffusion models (i.e., E-VDVAE~\cite{vae} and ADM~\cite{diffusion}, respectively).

DCGAN is an early model having a rather simple structure compared to the other models, thus its performance is generally considered to be worse than the others.
In the case of BigGAN, a smaller threshold value for the truncation trick yields a lower level of diversity of the generated images, which usually imposes a penalty on evaluation. 
StyleGAN2 has modified some layers of StyleGAN to improve the quality of generated images and to resolve blob artifacts of generated images.
Recently, ADM has shown remarkable performance with accessible likelihood measures, which is even better than GANs.
Although E-VDVAE can also measure the likelihood, generated images are often blurred.
Appropriate evaluation metrics should be able to identify these characteristics of the models.

\begin{table}[t]
\centering
\caption{Evaluation results of generative models using various metrics}
\label{tab:benchmark}
\setlength\tabcolsep{5pt}
\scriptsize
\begin{tabular}{llcccccc}
\toprule
Model               & Dataset  & Precision($\uparrow$) & Recall($\uparrow$) & IS($\uparrow$)    & FID($\downarrow$)   & KID($\downarrow$)   & TREND($\downarrow$)  \\
\midrule
DCGAN                       & CIFAR10  & 0.875     & 0.164  & 3.0   & 45.44 & 0.0316 & 0.0181 \\
\midrule
E-VDVAE & ImageNet & 0.568 & 0.439 & 6.9 & 46.98 & 0.0407 & 0.0275\\
BigGAN (0.2)    & ImageNet & 0.958     & 0.007  & 330.4 & 24.97 & 0.0118 & 0.0165 \\
BigGAN (0.4)    & ImageNet & 0.960     & 0.025  & 321.2 & 20.20 & 0.0106 & 0.0139 \\
BigGAN (0.6)    & ImageNet & 0.962     & 0.097  & 292.5 & 15.45 & 0.0078 & 0.0106 \\
BigGAN (0.8)    & ImageNet & 0.954     & 0.159  & 250.5 & 11.09 & 0.0048 & 0.0069 \\
BigGAN ($\infty$) & ImageNet & 0.916     & 0.294  & 144.6 & 6.24 & 0.0009  & 0.0032 \\
ADM-C & ImageNet & 0.872 & 0.669 & 61.8 & 8.91 & 0.0076 & 0.0043 \\
ADM-U & ImageNet & 0.890 & 0.705 & 208.6 & 4.94 & 0.0015 & 0.0026 \\
\midrule
E-VDVAE & FFHQ & 0.865 & 0.199 & 2.4 & 33.83 & 0.0230 & 0.0213 \\
StyleGAN                    & FFHQ     & 0.795     & 0.491  & 4.7   & 4.58 & 0.0011  & 0.0014 \\
StyleGAN2                   & FFHQ     & 0.768     & 0.585  & 4.8   & 3.09 & 0.0006  & 0.0012 \\
\bottomrule
\end{tabular}
\end{table}

In \tablename~\ref{tab:benchmark}, the evaluation results are shown.
In the case of IS, DCGAN, StyleGAN, and StyleGAN2 show extremely low scores compared to BigGAN, which indicates that IS is inadequate to evaluate GANs trained on datasets other than ImageNet.
Moreover, as the truncation threshold of BigGAN increases, IS rather decreases because IS does not consider the diversity of generated images.
In the case of the two-dimensional metric, precision yields inaccurate results, showing fluctuating scores with respect to the increase of the truncation threshold and judging that superiority of StyleGAN over StyleGAN2.
KID underrates the performance of diffusion models, by favoring BigGAN ($\infty$) over ADM-U.
Both TREND and FID show sensible evaluation results, while FID has weakness related to the number of test samples, which will be discussed in the next section.

\subsection{Robustness to the Number of Samples}
Since performance evaluation of GANs are based on sample statistics, robustness of an evaluation metric against the number of test data is highly desirable. Otherwise, the result of performance evaluation and comparison would change depending on the number of test data. Thus, when a robust metric is used, it is not restricted to generate as many images as in the original target dataset. Furthermore, a robust metric enables performance comparison across different studies that use different numbers of samples.
In this section, we demonstrate the robustness of the proposed method against the number of test data.

For the experiment, we randomly drop some of the generated image data to keep a certain proportion (i.e., 1, 1/5, 1/10, and 1/50), while the ground truth datasets remain the same. The generated images used in Section~\ref{sec:benchmark} are used.

\begin{figure}[t]
\centering
\subfloat[]{\includegraphics[width=0.43\linewidth]{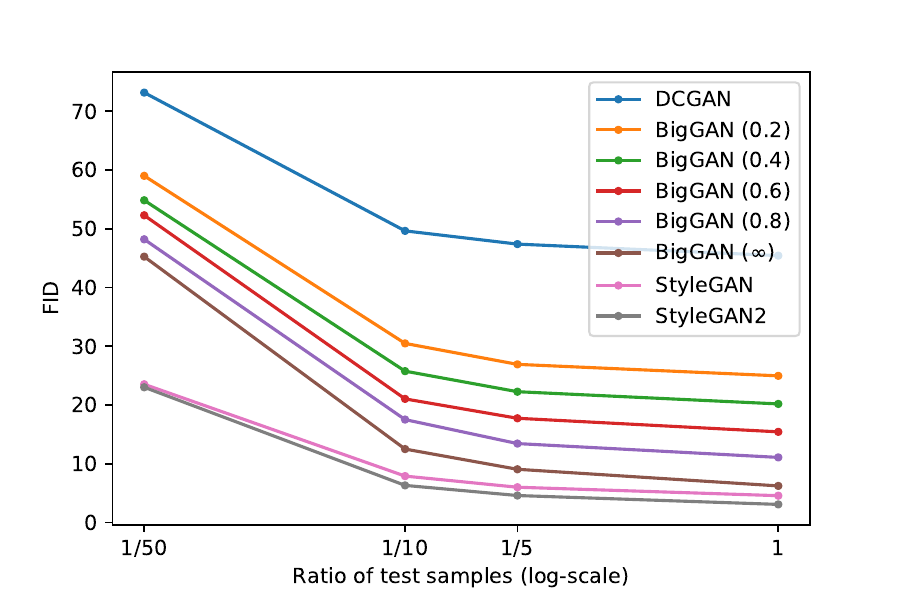}%
\label{fig:num1}
}
\hspace{1em}
\subfloat[]{\includegraphics[width=0.43\linewidth]{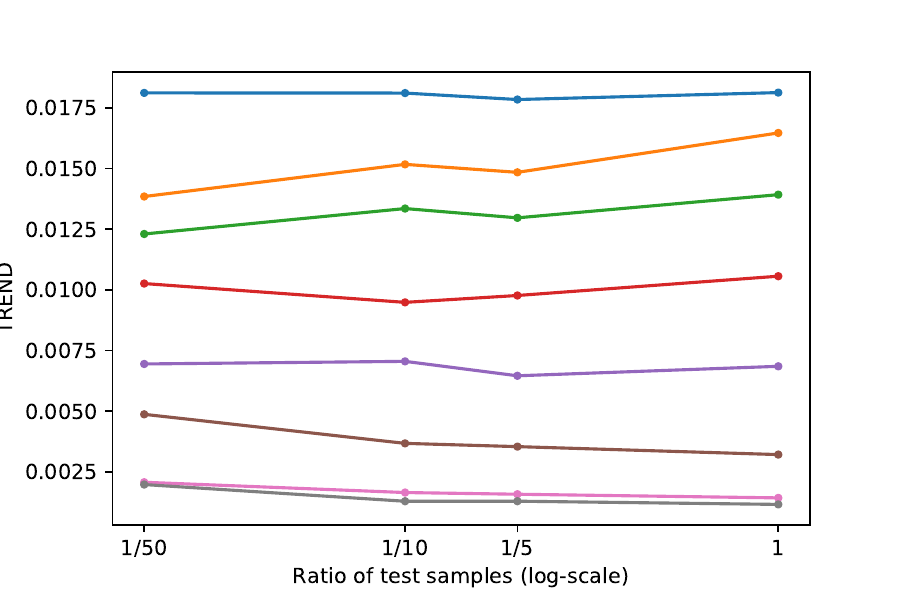}%
\label{fig:num2}}
\caption{Evaluation of DCGAN, BigGAN, StyleGAN, and StyleGAN2 with respect to the number of test samples using (a) FID and (b) TREND. The values in parentheses of BigGAN are the threshold values for the truncation trick and `$\infty$' means that the trick is not used.}
\label{fig:num}
\end{figure}

The results are shown in \figurename~\ref{fig:num}. 
FID in~\figurename~\ref{fig:num1} significantly varies with respect to the number of samples. In particular, it always decreases by increasing the number of samples.
Thus, using more samples may be wrongly interpreted as being better in image generation when FID is used.
On the other hand, TREND is hardly affected by the number of samples, resulting in consistent scores across different numbers of test data for all models in \figurename~\ref{fig:num2}.

\begin{figure}[t]
\centering
\subfloat[1 (50000 images)]{\includegraphics[width=0.36\linewidth]{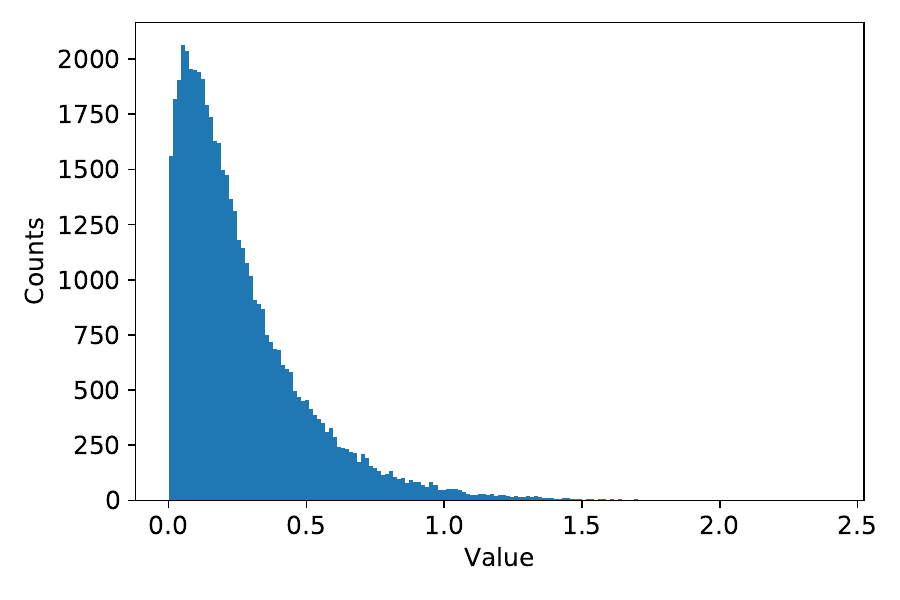}%
\label{fig:act}
}
\hspace{1em}
\subfloat[1/10 (5000 images)]{\includegraphics[width=0.36\linewidth]{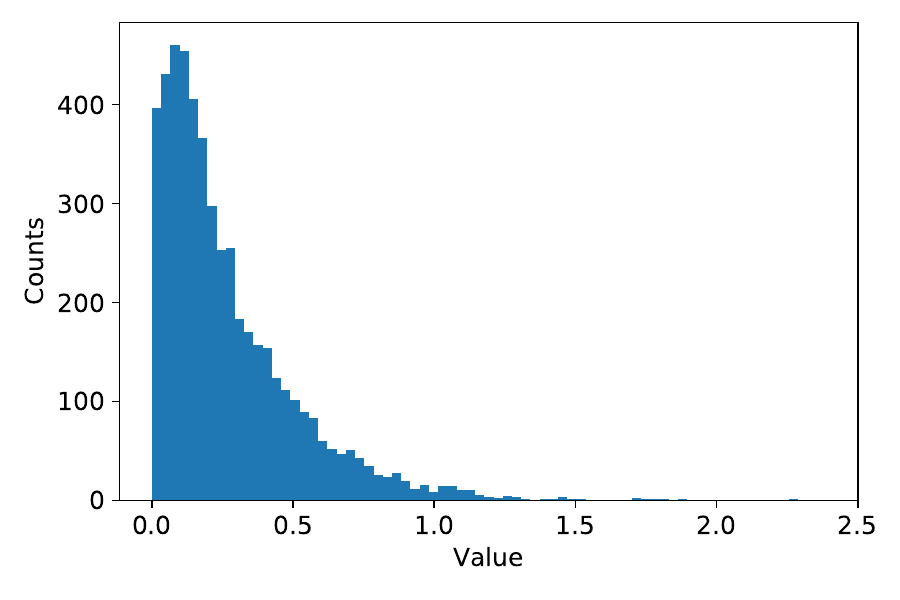}%
\label{fig:act10}}
\caption{Histograms of the Inception features for the 120th dimension with (a) 50000 and (b) 5000 images.}
\label{fig:num_act}
\end{figure}

FID is sensitive to the number of samples because its density estimation is inaccurate. As shown in \figurename~\ref{fig:num_act}, the distributions of the Inception features appear similar for different numbers of samples\footnote{The Mann-Whitney U test fails to reject the null hypothesis that the two distributions are statistically identical for 1992 (97.3\%) out of 2048 dimensions.}. Thus, accurate estimation of these distributions in TREND does not change much. In the case of FID, however, we observe that although the first term accounting for difference of $\mu$ in~\eqref{eq:FID} remains almost the same for different numbers of test data, the second term for difference of $\Sigma$ in~\eqref{eq:FID} increases when the number of data is reduced.

\begin{figure}[t]
\centering
\subfloat[\textit{threshold}=0.4]{\includegraphics[width=0.37\linewidth]{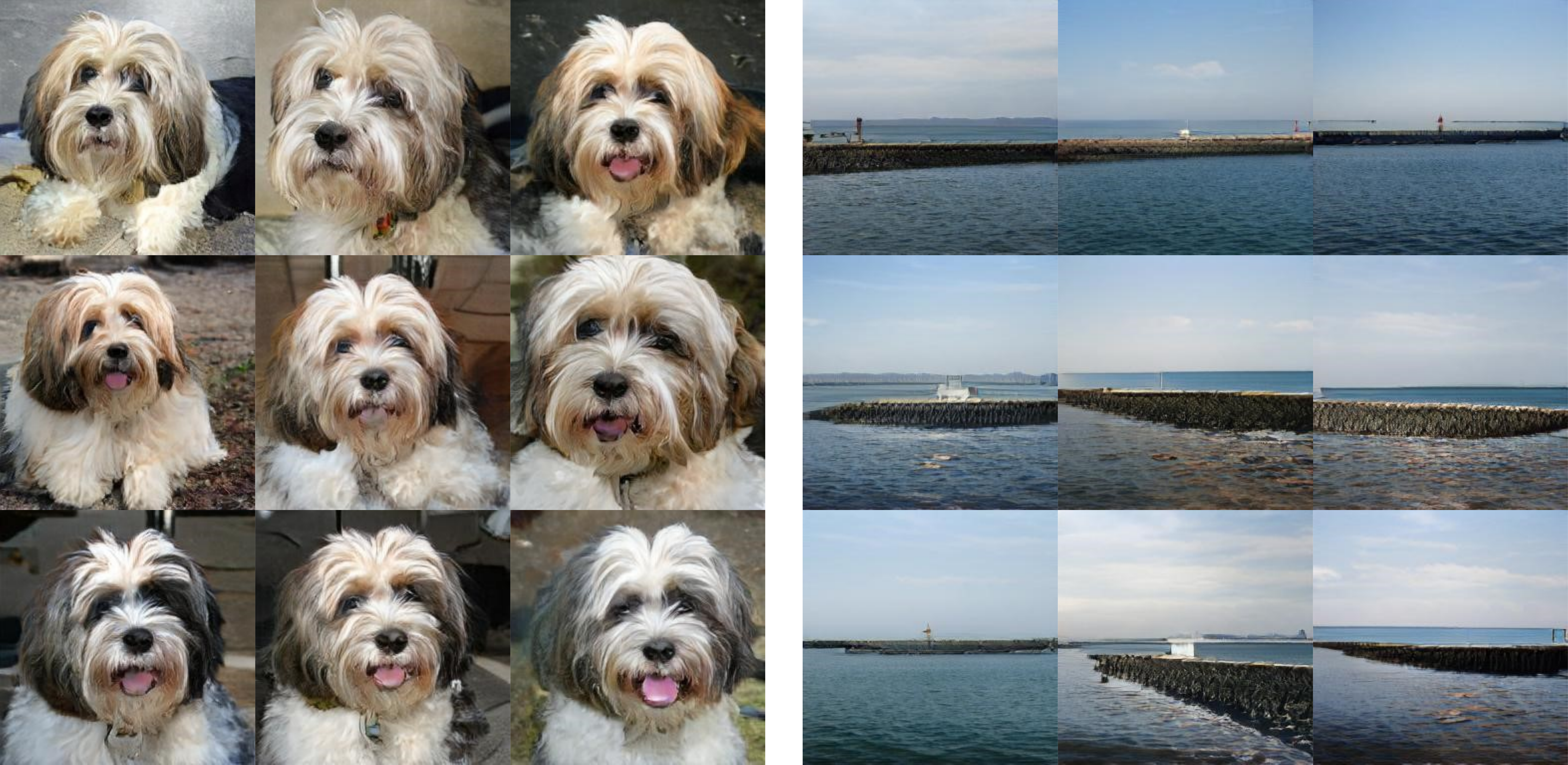}%
\label{fig:t4}
}
\hspace{1em}
\subfloat[\textit{threshold}=0.6]{\includegraphics[width=0.37\linewidth]{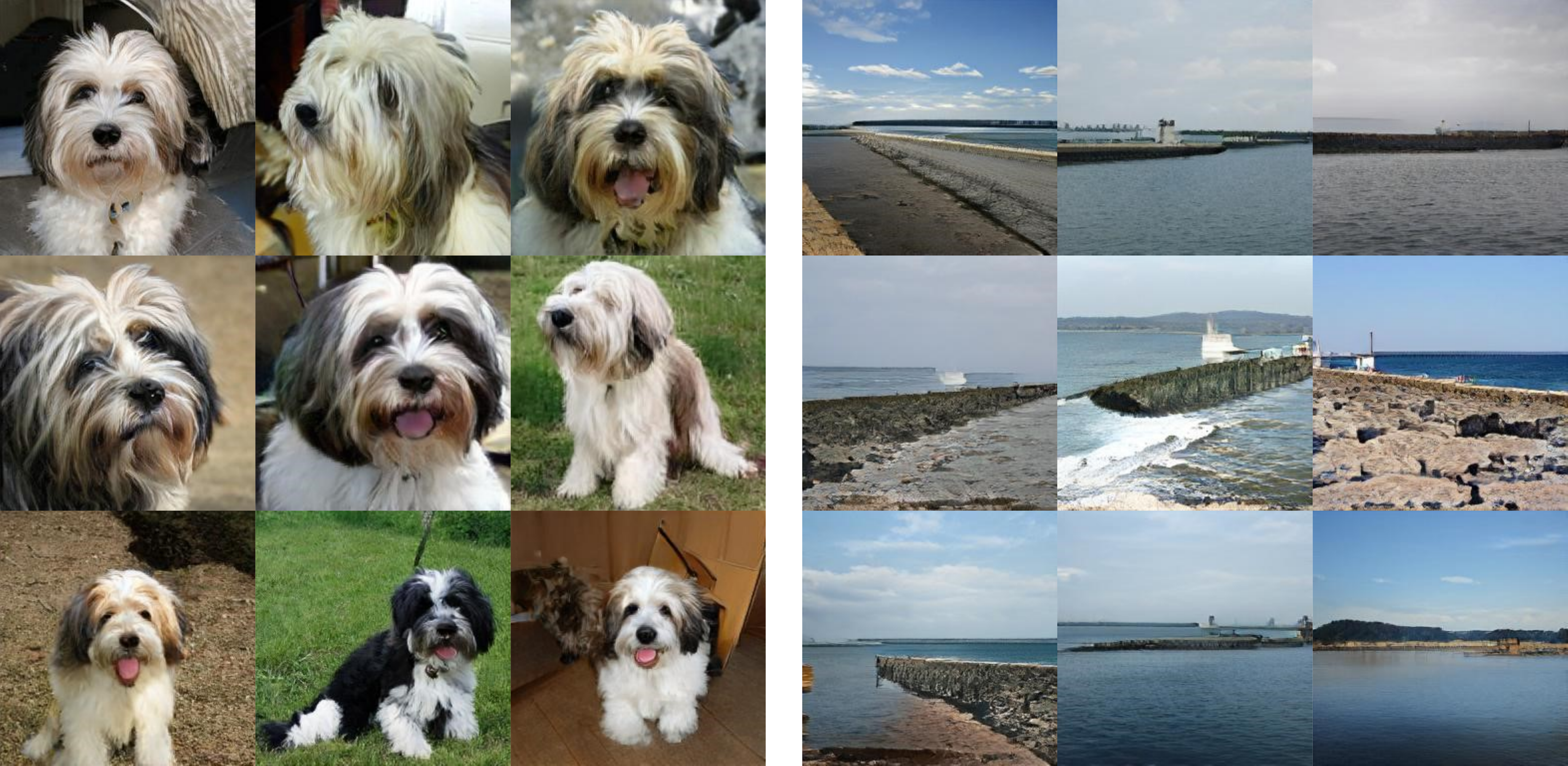}%
\label{fig:t6}}
\caption{Generated images of `Tibetan terrier' (left) and `breakwater' (right) from BigGAN with threshold values of (a) 0.4 and (b) 0.6 for the truncation trick.}
\label{fig:gen}
\end{figure}

\begin{table}[t]
  \centering
  \caption{Evaluation results using FID and TREND for different threshold values and numbers of samples. For a threshold of 0.6, we repeat sampling of 5000 images 10 times and the minimum, mean, and maximum values of FID and TREND are shown.}
  \label{tab:num}
  \setlength\tabcolsep{8pt}
  \scriptsize
  \begin{tabular}{ccc}
    \toprule
    Threshold (\# images)& FID & TREND  \\
    \midrule
    0.4 (50000) & \textbf{20.2} & 0.0139\\
    0.6 (5000) & 20.7/21.1/21.5 & \textbf{0.0091/0.0094/0.0098}\\
  \bottomrule
\end{tabular}
\end{table}

Due to the bias caused by the number of samples, evaluation of GANs using FID may be misleading. In \figurename~\ref{fig:gen}, generated images using BigGAN with different threshold values are shown. 
Overall, the quality of the generated images for both threshold values is similar. 
However, the images for the larger threshold value are more diverse (\figurename~\ref{fig:t6}), while the pose and background are almost identical in \figurename~\ref{fig:t4} due to the smaller threshold value. 
As a result, the model with a threshold of 0.6 is preferable to that with a threshold of 0.4. The evaluation results in~\tablename~\ref{tab:num} show that TREND is consistent with this observation, whereas FID is not due to the undesirable influence of the number of samples (\figurename~\ref{fig:num1}).

\section{Conclusion} \label{sec:conclusion}
We proposed a novel metric called TREND for evaluation of GANs.
We performed in-depth analysis of the Inception feature and showed the invalidity of the normality assumption used in the existing metrics. We used the truncated generalized normal distribution for more accurate density estimation of the Inception feature, based on which the proposed TREND was designed. The experimental results demonstrated that TREND is reliable in density estimation, effective for GAN evaluation, and robust against the number of samples.
In the future, we expect that TREND can be applied to various domains such as audio, motion pictures, or multimodal data, with proper selection of feature spaces.

\medskip{\smallskip \noindent \textbf{Acknowledgements.} This work was supported in part by the Artificial Intelligence Graduate School Program, Yonsei University under Grant 2020-0-01361, and in part by the Ministry of Trade, Industry and Energy (MOTIE) under Grant P0014268. }

\clearpage
%
%
\bibliographystyle{splncs04}
\bibliography{egbib}

\clearpage
\setcounter{equation}{0}
\setcounter{figure}{0}
\setcounter{table}{0}
\setcounter{section}{0}
\makeatletter
\renewcommand{\thesection}{\Alph{section}}
\renewcommand{\theequation}{\Alph{equation}}
\renewcommand{\thefigure}{\Alph{figure}}

\section{Inter-dimensional independence of Inception features} \label{sup:independence}
\begin{figure}[ht]
\centering
\includegraphics[width=0.5\linewidth]{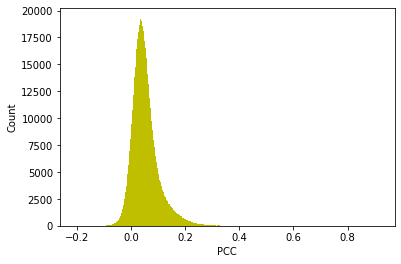}
\caption{Histogram of PCCs between pairs of Inception feature dimensions.}
\label{fig:pcchist}
\end{figure}

\begin{figure}[t]
\centering
\includegraphics[width=0.6\linewidth]{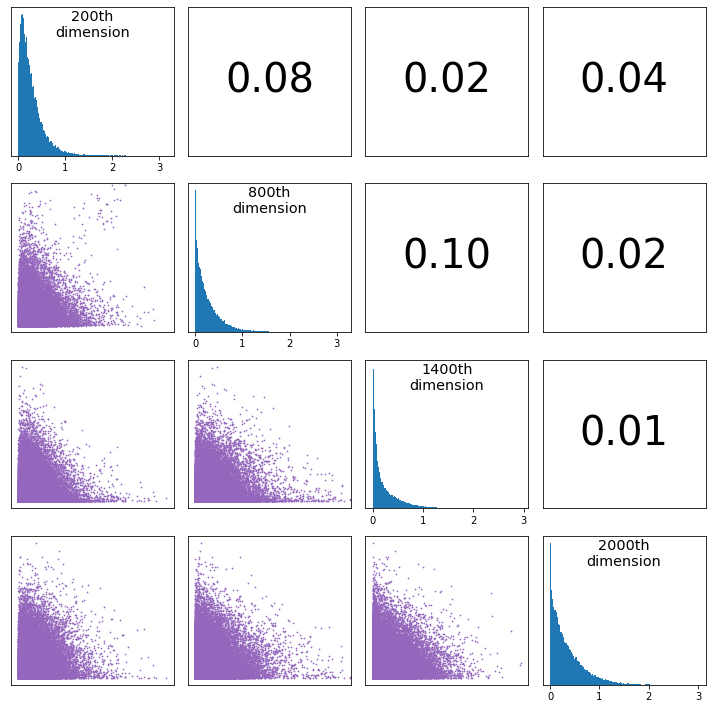}
\caption{Scatter plots showing the inter-dimensional relationships of Inception features for selected feature dimensions (200th, 800th, 1400th, and 2000th). The corresponding PCC values are also shown. The histogram of each dimension is shown on the diagonal.}
\label{fig:scatter}
\end{figure}

\figurename~\ref{fig:pcchist} shows the histogram of the PCC value between each pair of dimensions of Inception features, which shows near independence between feature dimensions (Section~\ref{sec:analysis}).
The scatter plots in~\figurename~\ref{fig:scatter} also confirm approximate independence between dimensions.

\section{Experimental Setup} \label{sup:setup}
For target images, we use the CIFAR10~\cite{cifar10}, CelebA~\cite{celeba}, Flicker-Face-HQ (FFHQ)~\cite{sg}, and ImageNet~\cite{imagenet} datasets. We use the validation split of the datasets or the test split only if the validation split is not provided. 
For evaluation of generative models, we generate images using the pre-trained models as follows: DCGAN~\cite{dcgan} trained on the CIFAR10 dataset, ProGAN~\cite{progan} trained on the CelebA dataset, BigGAN-deep-256~\cite{biggan}, ADM~\cite{diffusion}, E-VDVAE~\cite{vae}, trained on the ImageNet dataset, and StyleGAN~\cite{sg}, StyleGAN2~\cite{sg2}, and E-VDVAE~\cite{vae} trained on the FFHQ dataset.

We use the pre-trained DCGAN~\cite{dcgan} model\footnote{https://github.com/csinva/gan-vae-pretrained-pytorch} trained on CIFAR10 to generate 32$\times$32 images. 
We generate 256$\times$256 images using the pre-trained BigGAN~\cite{biggan} model\footnote{https://tfhub.dev/deepmind/biggan-deep-256/1} trained on ImageNet. 
Since BigGAN is trained on the conditional class label of ImageNet, we generate a fixed number (50) of images for each class. We also apply the truncation trick for BigGAN with varying the threshold value of latent vectors.
In general, a smaller threshold value yields a lower level of diversity of the generated images. 
We use five different threshold values: 0.2, 0.4, 0.6, 0.8, and $\infty$ (i.e., no truncation trick).
We use StyleGAN~\cite{sg} that is trained on FFHQ to generate high-resolution face images having a resolution of 1024$\times$1024 pixels. 
We use samples downloaded from its official website\footnote{https://github.com/NVlabs/stylegan}. 
For StyleGAN2, We use the pre-trained model\footnote{https://github.com/NVlabs/stylegan2-ada-pytorch}~\cite{sg2} that is also trained on the FFHQ dataset.
We use the pre-trained ADM-C and ADM-U~\cite{dcgan} models\footnote{https://github.com/openai/guided-diffusion} trained on ImageNet with classifier guidance and up-sampling, respectively.
For E-VDVAE~\cite{vae}, we use the pre-trained models\footnote{https://github.com/Rayhane-mamah/Efficient-VDVAE} trained on ImageNet and FFHQ datasets.

The initial values of the parameters in our method are so determined that fast convergence of the maximum likelihood estimation using~(10) is achieved. The initial value of $\mu$ is set to the peak location of the histogram of Inception features for each dimension. The initial values of $\sigma$ and $\beta$ are empirically set to $1.5\hat{\sigma}$ and 0.67, respectively, where $\hat{\sigma}$ is the sample standard deviation.
It takes about 1.5 hours to estimate the distributions of 2048-dimensional Inception features for 50000 images using a \SI[product-units=single]{3.7}{\giga\hertz} quad-core Intel Xeon\textregistered~ CPU.

\section{More Results of Density Estimation} \label{sup:realdensity}
\begin{figure}[t]
\centering
\subfloat[CIFAR10]{\includegraphics[width=0.75\linewidth]{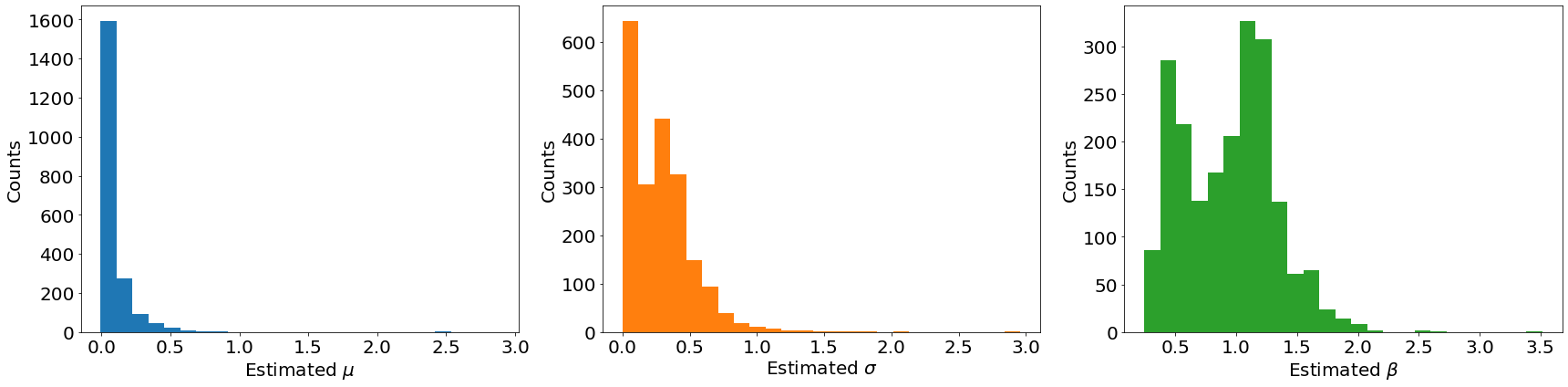}%
\label{fig:hp_cifar}
}
\hfill
\subfloat[CelebA]{\includegraphics[width=0.75\linewidth]{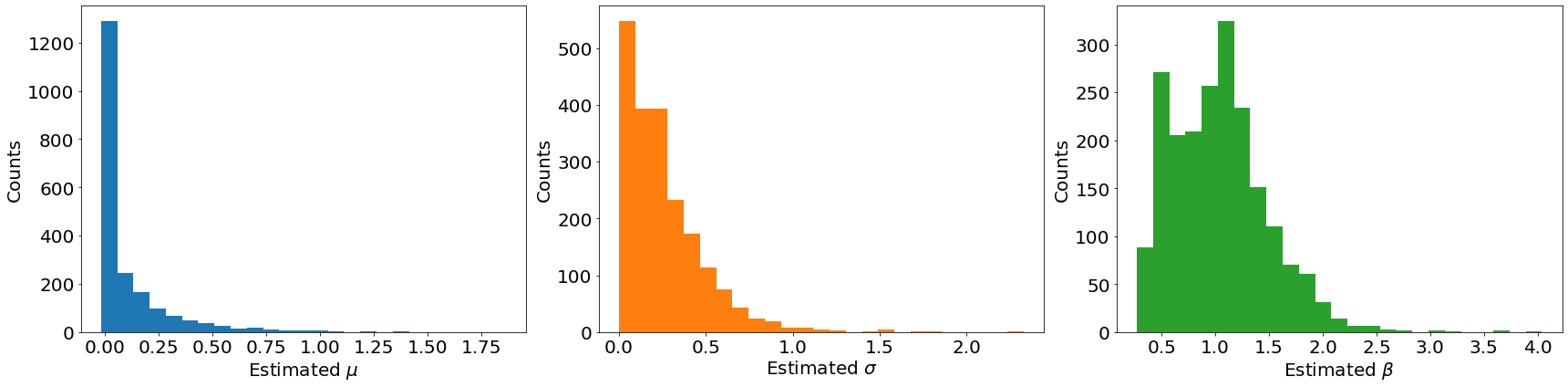}%
\label{fig:hp_celeba}}
\hfill
\subfloat[ImageNet]{\includegraphics[width=0.75\linewidth]{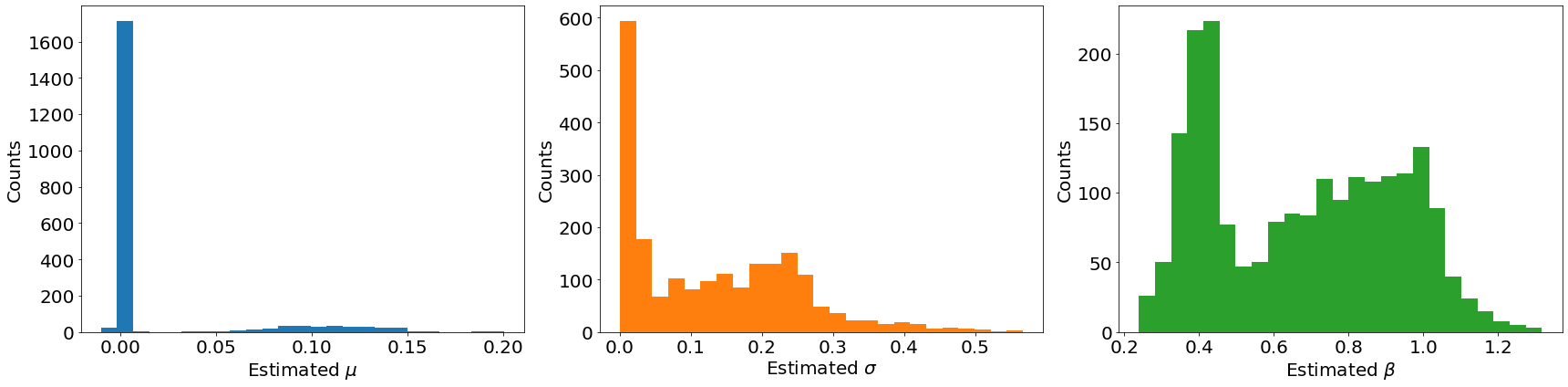}%
\label{fig:hp_imagenet}}
\hfill
\subfloat[FFHQ]{\includegraphics[width=0.75\linewidth]{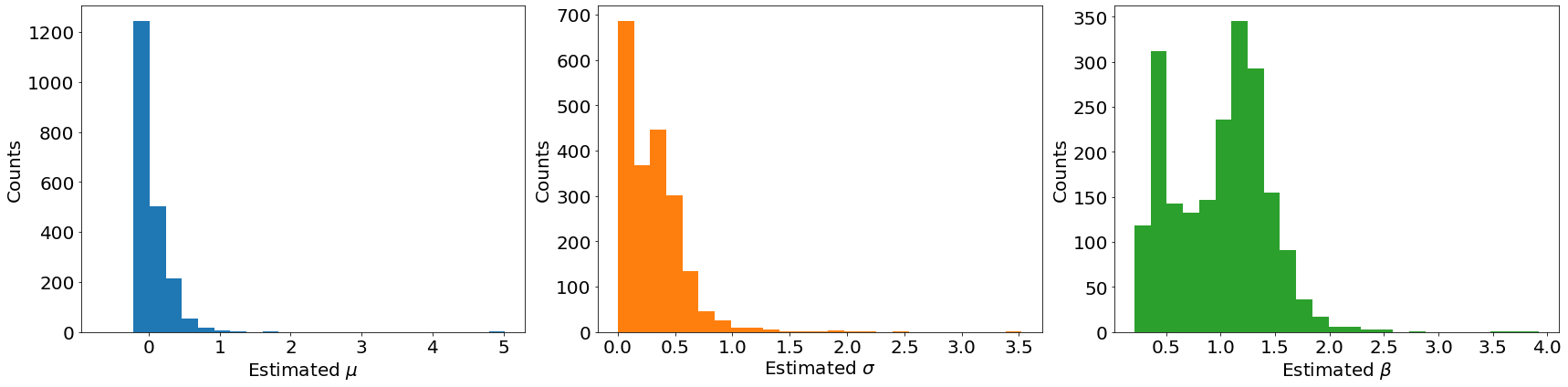}%
\label{fig:hp_ffhq}}
\caption{Histograms of the estimated parameters $(\mu,\sigma,\beta)$ for the 2048 dimensions of the Inception features for images from (a) CIFAR10, (b) CelebA, (c) ImageNet, and (d) FFHQ datasets.}
\label{fig:hp}
\end{figure}

To supplement the results in Section~\ref{sec:realdensity}, \figurename~\ref{fig:hp} shows the histograms of the parameters for all feature dimensions.

\section{Disturbed Images} \label{sup:disturbance}
\begin{figure}[t]
\centering
\includegraphics[width=0.8\linewidth]{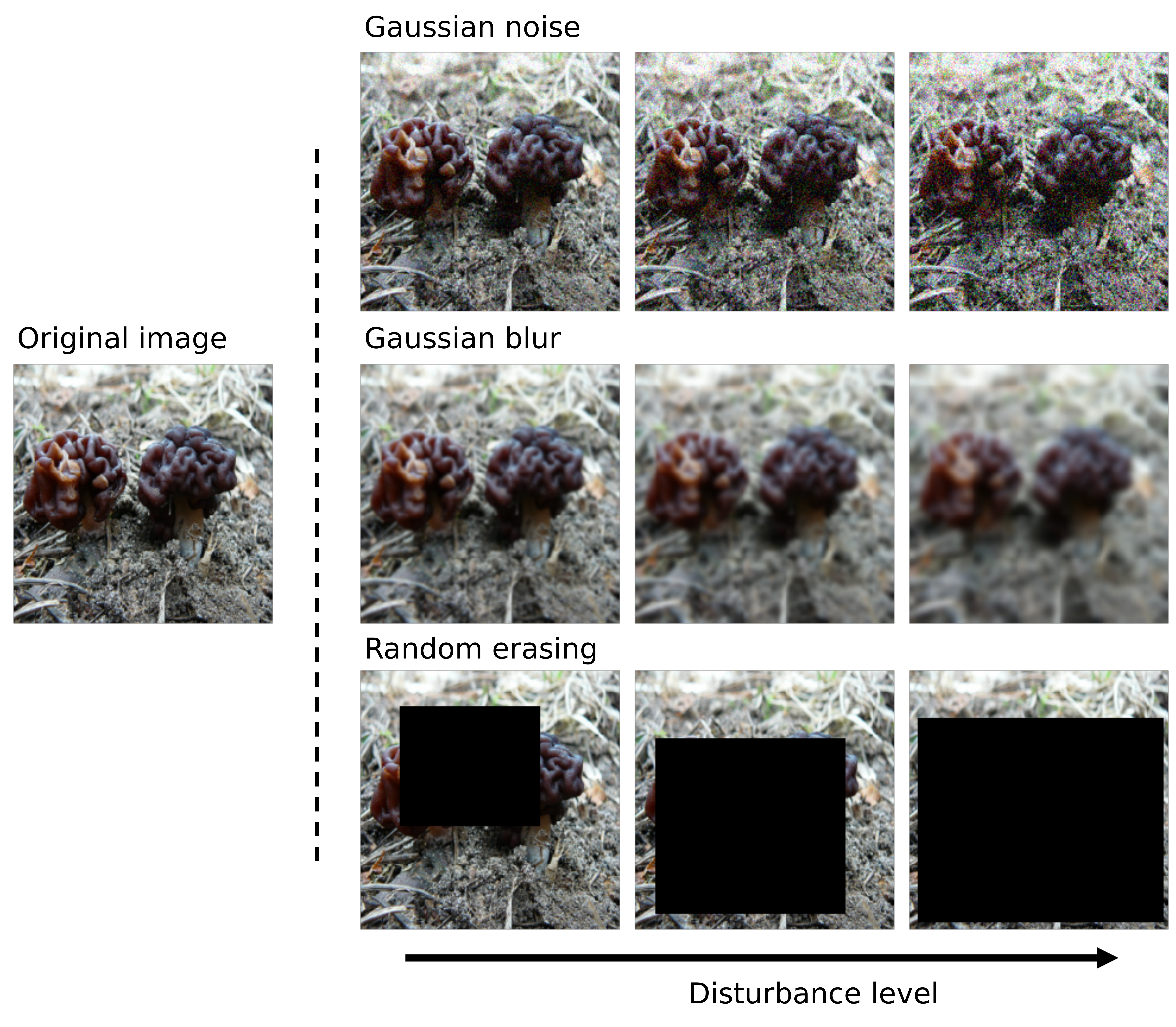}
\caption{Example of disturbed images.}
\label{fig:manipulation}
\end{figure}
An example of disturbed images used in Section~\ref{sec:disturbance} is shown in \figurename~\ref{fig:manipulation}.

\end{document}